%% file: TofighiRoMoVa_RAM26_final.tex
\documentclass[journal]{IEEEtran} 
\usepackage[english]{babel}
\usepackage{microtype}
\usepackage{graphicx}
\usepackage{tabularx}
\usepackage{subfigure}
\usepackage{booktabs} 
\usepackage{amsmath}
\usepackage{graphicx}
\usepackage{lipsum} 
\usepackage{amssymb}
\usepackage{tabularx}
\usepackage{pifont}
\usepackage{hyperref}

\usepackage{array}
\newcolumntype{P}[1]{>{\centering\arraybackslash}p{#1}}
\usepackage{color,colortbl} 

\begin{document}

\title{A Survey on Event-based Optical Marker Systems} 

\author{Nafiseh Jabbari Tofighi,\, Maxime Robic$^{\dag}$,\, Fabio Morbidi$^{\ast}$,\, Pascal Vasseur
\thanks{The authors are with the MIS laboratory, University of Picardie Jules Verne, 33 Rue Saint-Leu, 80039 Amiens, France.
Emails: \tt\small \{nafiseh.jabbari, maxime.robic, fabio.morbidi, pascal.vasseur\}@u-picardie.fr} 
\thanks{$^{\dag}$Current affiliation of M. Robic: DART Lab, Politecnico di Milano, Italy.}
\thanks{$^{\ast}$Corresponding author.}}

\markboth{IEEE Robotics \& Automation Magazine, Final Version, \today}  
{Jabbari Tofighi \MakeLowercase{\emph{et al.}}}

\maketitle 

\begin{abstract}
The advent of event-based cameras, with their low latency, high dynamic range, and reduced power consumption, marked a turning point in machine perception and robotic vision. In~particular, the combination of these neuromorphic sensors with widely-available passive or active optical markers (e.g. AprilTags, arrays of blinking LEDs), has recently opened up a new field of opportunities. This survey paper provides a comprehensive review of \emph{Event-Based Optical Marker Systems} (EBOMS). We~analyze the underlying principles and technologies on which these systems are based, with a special focus on their asynchronous operation and robustness against challenging lighting conditions. We also describe the most relevant applications of EBOMS, including object detection and tracking, pose estimation, and optical communication. The article concludes with a discussion of possible future research directions in this rapidly-emerging and multidisciplinary area. 
\end{abstract}


\IEEEpeerreviewmaketitle


\vspace{-0.2cm}
\section{Introduction}

\begin{figure}[t!]
    \centering
    \!\!\!\!\includegraphics[width=1.06\linewidth]{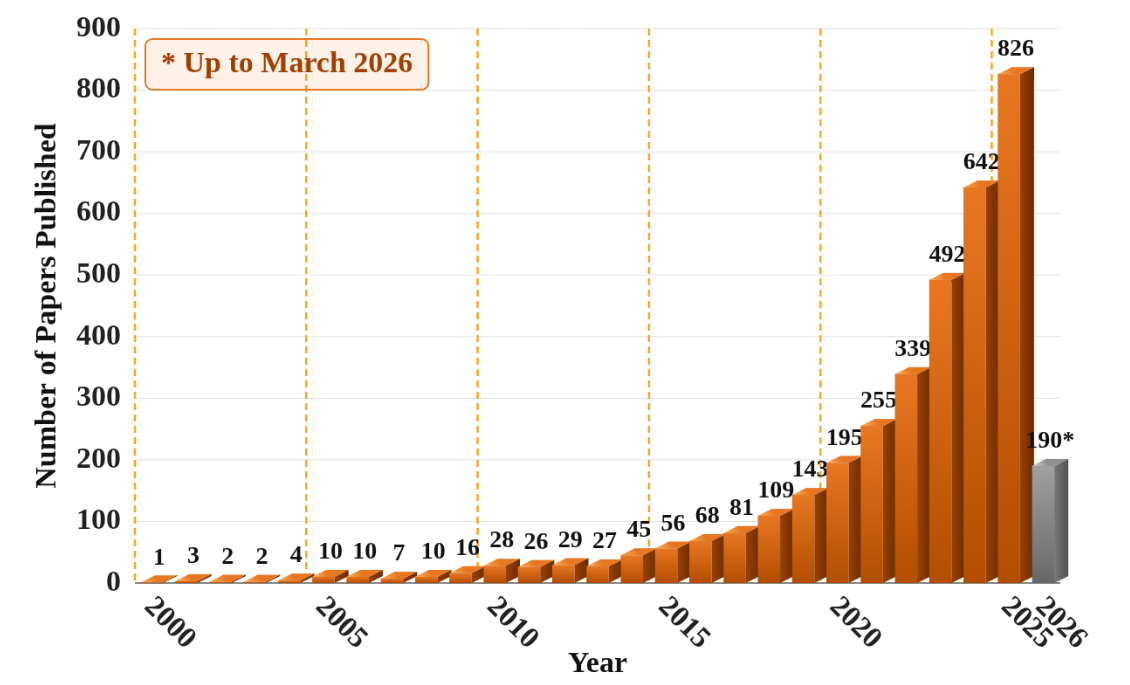}
    \vspace{-0.35cm}
    \caption{Number of publications on event cameras per year, based on~Scopus~\cite{Scopus_web} (data collected in March 2026).}
    \label{fig:event_camera_trend}
\end{figure}

\IEEEPARstart{I}{n} the last decade, there has been a noticeable trend towards the adoption of \emph{event-based sensors} in far-reaching engineering applications (see Fig.~\ref{fig:event_camera_trend}). The recent advances in neuromorphic sensing opened up entirely new ways of perceiving the world around~us, leading to a major paradigm shift in computer vision and image processing~\cite{gehrig2024low}. Regardless of the amount of motion or light changes within a scene, traditional frame-based cameras capture images at predefined, constant intervals. This operating mode is problematic when dealing with fast motions and variable light conditions in dynamic environments. In contrast, event cameras, inspired by the human visual system, capture light-intensity changes in an asynchronous fashion, minimizing data redundancy and energy consumption. Every brightness change within a scene triggers an event that records its spatial location (pixel coordinates), timestamp, and polarity~\cite{gallego2020event}. Event cameras have low latency with response times in the microsecond range, and their high dynamic range ensures effective operation in both low-light indoor and bright outdoor environments. All these features make event cameras a valid alternative to conventional vision sensors in a host of real-world applications~\cite{beck2021extended}. 

The ability of event cameras to capture rapid changes with minimal latency makes them particularly well-suited for mobile robotics~\cite{aitsam2024event,tenzin2024application,delbruck2016neuromorphic,cao2024eventboost,sanyal2024ev} and autonomous driving~\cite{chen2019event,li2019event,shariff2024event} applications, where traditional cameras suffer from motion blur and slow response times~\cite{dai2023hybrid}. Moreover, their temporal resolution encourages their use for tasks requiring high-speed visual feedback, such as tracking~\cite{daco_25, wang2024event}, pose estimation~\cite{da2025gyrevento, ren2024simple, yuan2024event}, and optical communication~\cite{wang2022smart}. As research progresses, event cameras have also continued to demonstrate their value in a wide range of computer-vision problems: motion segmentation~\cite{georgoulis2024out}, optical flow~\cite{gehrig2024dense,wan2024event} and depth~\cite{furmonas2022analytical} estimation, and image reconstruction~\cite{scheerlinck2020fast}. 

One of the most promising areas of application of event cameras, where their potential can be fully realized, is \emph{Event-Based Optical Marker Systems} (EBOMS\footnote{Throughout this paper, the acronym EBOMS is indifferently used in the singular or~plural.}). Optical markers are visual reference points which have been configured to encode or transmit data, and which can be easily detected by an event-based sensor~\cite{toyoura2014mono}. EBOMS take advantage of the low latency and high temporal resolution of event cameras for integrated communication, tracking and localization. In~contrast to traditional optical markers that rely on frame-based detection, event-based systems can effectively track blinking markers, irrespective of the ambient light conditions. The energy efficiency and scalability of EBOMS make them appropriate for use in edge devices~\cite{sridharan2024ev}, multi-sensor networks or large-scale autonomous systems~\cite{Wang_etal_ACMCS26}. 
Moreover, while traditional communication and tracking systems (e.g. radio-frequency~\cite{pandey2024uav} or Li-Fi communication~\cite{alfattani2021review}, and RGB camera-based tracking~\cite{hasan2021optical}) are sensitive to external perturbations, EBOMS are intrinsically robust. 

EBOMS rely on a broad range of optical markers, which are tailored to the specific application and operational requirements. LED-based active markers emit modulated light signals which can be processed by an event camera for rapid object tracking and communication~\cite{li2024review}. Geometric and passive markers (such as AprilTag or ArUco), can be exploited for accurate pose estimation and tracking as well. These~markers can be combined to give rise to hybrid systems, offering increased robustness and flexibility in complex environments. 

Despite their favorable reception, a~comprehensive review of the current landscape of EBOMS is still missing. In fact, the rapid progress of neuromorphic perception yielded a large body of research, fragmented across multiple scientific domains (Fig.~\ref{fig:event_camera_trend} shows that the number of publications on event cameras, has increased exponentially in the last 25 years). A~systematic review of existing methodologies, technologies, and applications in robotics, intelligent vehicles, augmented reality (AR), and telecommunications, is thus sorely needed. This paper tries to fill this gap by providing a comparative analysis of EBOMS in the last 15~years (it covers research between 2011 and~2026). By~identifying current trends and open challenges, this survey is intended to foster research and innovation in this highly-dynamic~field. 

In summary, the~original contributions of this paper can be stated as~follows:
\begin{enumerate}
    \item We conduct the first structured review of EBOMS, consolidating knowledge from event-based vision and optical-marker systems, and laying the groundwork for future research. The proposed taxonomy provides researchers with a clear understanding of theoretical underpinnings, roadblocks, and technical evolution over~time.  
    \item We analyze and critically compare the existing EBOMS according to different criteria: sensing technology, marker design, coding/decoding schemes, data processing and object tracking, and deployment in the field (see~Table~\ref{Table:Survey}).
    \item We review key applications of EBOMS, including object tracking, pose estimation, and optical communication, and we weigh up their pros and cons compared to the conventional frame-based systems.
\end{enumerate}
The remainder of this paper is organized as follows. Sect.~\ref{Sect:BackGround} provides background information on event cameras and optical-marker systems. In Sect.~\ref{Sect:Applications}, we give an account of the main applications of EBOMS. Finally, in Sect.~\ref{Sect:FutureWork}, the conclusions are drawn and several promising directions for future research are~discussed. 


\section{Background}\label{Sect:BackGround}

\begin{figure}[t!]
    \centering
    \includegraphics[width=1\linewidth]{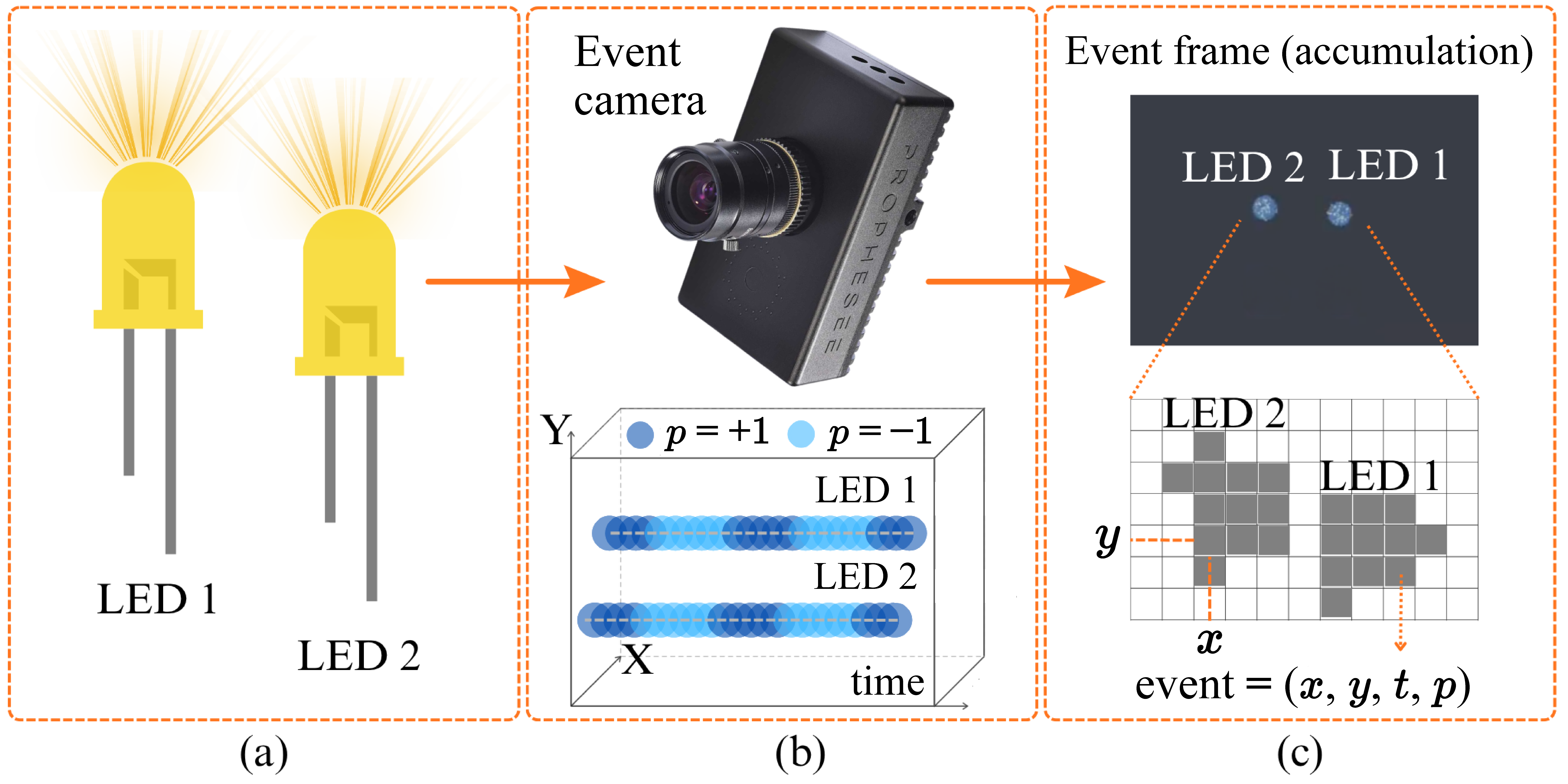}
    \vspace{-0.45cm}
    \caption{Core components of Event-Based Optical Marker Systems (EBOMS): (a) Optical-marker system, which may include different types of markers (e.g. LEDs) depending on the specific application; (b),(c) Event-based camera, where each pixel operates independently and generates events asynchronously in response to brightness changes, encoding spatial, temporal and polarity~information.}\label{fig: overview}
\end{figure}

This section briefly reviews the fundamental principles behind EBOMS, focusing on the key components and underlying technologies (see Fig.~\ref{fig: overview}). First, the event cameras are presented, highlighting their asynchronous operation and advantages over traditional frame-based sensors. Second, the existing optical-marker systems are categorized, and their role in visual tracking, localization, and communication, is explained. 

\subsection{Event-based cameras}

Event (or event-based) cameras are bio-inspired sensors which detect brightness changes in a scene, instead of capturing frames at fixed rates. Each pixel in an event camera operates independently, and it is sensitive to light-intensity fluctuations occurring in the environment. An event can be represented by the quadruple $(x,\, y,\, t,\, p)$, where $(x,\,y)$ are the pixel \emph{coordinates} of the event, $t$ is the \emph{timestamp}, and $p = \pm 1$ is the \emph{polarity} of the event. If $p$ is positive (negative), it means that the brightness has increased (decreased) at pixel $(x,\, y)$ since the last event~\cite{gallego2020event}.

Among the distinctive advantages offered by the sensing mechanism of event cameras, are asynchronous operation and high temporal resolution. With a response time in the microsecond range, these cameras are well equipped for sensitive applications requiring high accuracy and fast processing times. The high dynamic range is another strength of event cameras, especially in outdoor environments with highly-variable or low-light conditions, where conventional frame-based cameras usually struggle. Finally, the event-based data collection mechanism, by only taking intensity variations into account and by disregarding those parts of the scene that do not change over time, guarantees high energy efficiency, which is crucial in applications involving portable battery-powered~devices.

\begin{figure*}[t!]
    \centering
    \includegraphics[width=0.9\linewidth]{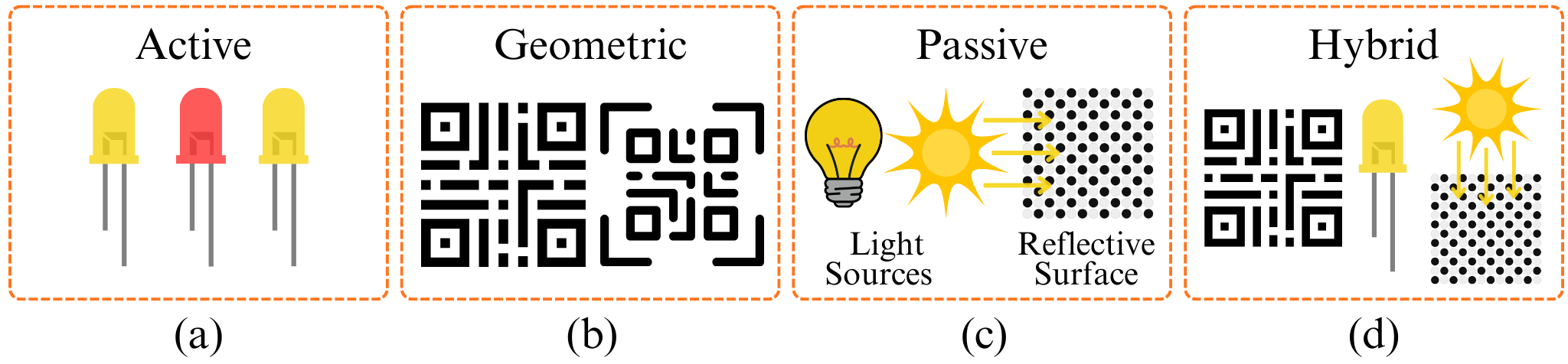}
    \vspace{-0.2cm}
    \caption{Classification of optical markers used in EBOMS: (a) Active markers emit modulated light signals, typically using LEDs; (b)~Geometric markers, such as AprilTag or ArUco, encode information through distinct visual patterns; (c) Passive markers depend on ambient or external light sources; (d) Hybrid markers combine active, geometric, or passive elements to capitalize on their respective strengths.} \label{fig: markers} 
\end{figure*}

\input{Tables/comparison_EBOMS-FBcameras.tex}

\input{Tables/survey_papers.tex}

\subsection{Optical-marker systems}

The use of optical markers for vision-based tracking and localization, and data encoding has a long history. The existing systems rely on visible or infrared (IR) lights that act as reference points for the receiver. Visual fiducials can be categorized as either \emph{active} or \emph{passive}. Active markers emit their own light and require a power source, such as light emitting diodes (LEDs), while passive markers rely on reflected light from external sources. Independent of this classification, many fiducials are designed with distinct geometric patterns, such as AprilTag, ARTag, ArUco, or QR codes. 
These geometric markers can be implemented either actively (for example, using computer screens) or passively (using printed patterns). For specific applications, active and passive markers can be combined into hybrid systems, which often provide higher accuracy, improved robustness, and better~adaptability to changing environmental conditions.

EBOMS go beyond frame-based systems and take advantage of flashing optical markers. Traditional frame-based optical communication systems are fundamentally limited by their frame rate, which typically ranges between tens and hundreds of frames per second (fps), with high-speed cameras reaching into the thousands of fps. This restricts their transmission rate to the kilobit-to-megabit per second range, depending on the specifications of the camera~\cite{cahyadi2020optical}. Other types of fast optical receivers, such as photodiodes, can operate at gigahertz frequencies: however, they function as single-point sensors and have poor spatial resolution, making them unsuitable for vision-based detection and communication tasks~\cite{nishar2025non}. In contrast, EBOMS operate asynchronously, with bandwidths ranging from several megahertz to beyond a gigahertz, enabling transmission rates in the gigabit-per-second range~\cite{gallego2020event}. The~seamless fit between the high temporal resolution of event cameras, which selectively capture only rapid visual changes, and the capability of LEDs to operate at high frequency, play a pivotal role in the superior performance of EBOMS in numerous tasks. Table~\ref{Table:Comparison}~provides a comparative overview of EBOMS and conventional frame-based vision systems, across key performance criteria. Overall, EBOMS offer clear advantages in terms of latency, dynamic range, robustness to motion blur, and energy efficiency, while~spatial resolution and technical maturity are two prerogatives of frame-based systems.  


\section{Applications}\label{Sect:Applications}

Because of their low latency and resilience under extreme conditions, EBOMS are being used in a number of classical and emerging applications. Although, in principle, EBOMS could be categorized according to the type of application (for~instance, tracking, communication or pose estimation), the lines between them are often blurry. Some works have developed integrated frameworks that simultaneously alleviate the need for tracking or communication, or utilize pose estimation within a joint localization and navigation pipeline. This convergence demonstrates the adaptability of EBOMS, since the same technology can be used for a variety of purposes.

Table~\ref{Table:Survey} presents an overview of existing EBOMS, classified according to different axes (camera/marker technology, operating conditions, frequency, etc.). In the rest of this section, we~describe the systems reported in Table~\ref{Table:Survey} in more detail. We start our exposition by examining works that are primarily geared towards detection and tracking of optical markers, as this serves as the basis for many approaches in this domain. The~pose-estimation problem is then addressed, and finally, EBOMS for communication tasks are discussed. As reported in Fig.~\ref{fig: markers}, EBOMS utilize four classes of optical markers: Sect.~\ref{SectApp-comm} is accordingly divided into four parts, covering the case of \emph{active}, \emph{geometric}, and \emph{passive markers}, and of \emph{hybrid~systems}.

\subsection{Object detection and tracking}\label{SectApp-ODT}

Detecting and tracking a moving target is a fundamental task in many engineering applications, including robotics and autonomous vehicles~\cite{kanellakis2017survey}. Traditional frame-based systems typically handle this task, although their slow response times introduce delays~\cite{bai2022infrastructure,iaboni2021event}. This constraint makes them unsuitable for time-sensitive applications. Furthermore, classical detectors and trackers often rely on the appearance of the entire target (for example, the airframe of a drone), which increases the likelihood of false detections and requires a substantial computational budget. A more efficient strategy is to detect only specific elements attached to the target, such as ArUco or active optical markers. This reduces the search space to a~narrow region inside the scene: moreover, the marker itself can convey additional information useful for pose estimation or communication.

One of the earliest works exploring this idea is~\cite{muller2011miniature}, where the authors used event-based sensors to overcome the limitations of conventional tracking methods. They proposed two algorithms for real-time 2D tracking of blinking LEDs. The first algorithm accumulates events at each pixel over a fixed time window and the LED's position corresponds the pixel with the highest event count. While simple and effective, this method relies on a window of fixed size. To counter this shortcoming, a second algorithm is introduced, which asynchronously follows events at each pixel and identifies patterns matching the LEDs' frequency. The effectiveness of these methods has been demonstrated for autonomous robot homing and high-speed pan-tilt tracking, laying the foundation for subsequent EBOMS research.

Geometric markers have also been considered for fast tracking and orientation estimation. In~\cite{loch2023event}, ArUco markers were used for continuous high-speed tracking, reaching 156~kHz with end-to-end latency as low as 3~ms. The method begins with a polarity-map-based ArUco detection, then updates the pose using the known marker geometry, and finally performs a two-way verification step to detect tracking failures by comparing forward and backward pose predictions.

Beyond the algorithms themselves, marker design itself can significantly influence EBOMS' performance. In~\cite{zhang2023improved}, the authors introduced active ArUco markers incorporating multiple LEDs, enabling reliable detection in scenarios with fast motions, low light, or high contrast. Their pipeline integrates mean-shift filtering for noise reduction, a geometric corner-refinement stage, and a Hamming-distance-based error-detection scheme. Adaptive thresholding with weighted statistics improves robustness under motion blur and partial occlusions. The results in~\cite{zhang2023improved}, show clear advantages over traditional printed markers. Along similar lines, the authors in~\cite{huang2023novel} presented EVNT-ArUco, a blinking ArUco marker specifically designed for event cameras, offering improved performance in static and low-light environments, compared to passive markers.

Event-based tracking methods leverage asynchronous sensing to overcome the limitations of frame-based systems, with the early LED-based approaches demonstrating feasibility, and the more recent ArUco-based methods, increasing speed and robustness. Marker design plays a central role, and both active and blinking geometric markers have shown significant improvements in reliability, especially under challenging lighting and motion conditions.

\subsection{Pose estimation}\label{SectApp-PE}

EBOMS hold tremendous potential for pose estimation in a wide range of scenarios, including portable motion capture systems and augmented and virtual reality. Accurate and real-time pose estimation is crucial in these applications, as it offers the spatial information required for object tracking or trajectory planning in both physical and virtual environments. However, conventional optical systems use frame-based cameras which capture scenes at fixed rates, leading to motion blur during rapid motions and performance degradation in low-light or high-contrast conditions~\cite{cassinis2019review}. This might result in erroneous feature detections or loss of visual correspondences between frames. As a consequence, the estimated pose may drift over time or become unstable in the presence of sharp accelerations, which negatively affects tracking accuracy in real-world conditions. To get around these well-known limitations, a possible option is to combine event cameras with optical markers. 

A first step in this direction was taken in~\cite{xu2019method}, providing a baseline for subsequent studies. The proposed system consists of four LEDs, whose high-frequency flashes are captured by an event cameras. After a preprocessing step, a low-pass filter removes slow variations, followed by a clustering step to detect isolated LED events. The initial camera pose is then estimated using the Perspective-$n$-Point (PnP) algorithm, with an additional optimization step to minimize the reprojection error. Finally, the 6-DoF estimates are refined using a graph-based optimization method.

Building upon this blueprint, the authors in~\cite{chen2020novel} proposed an indoor positioning system based on an event camera and multiple LEDs. Each LED is assigned a unique frequency, detected through a Gaussian Mixture Probability Hypothesis Density (GM-PHD) filter for tracking, combined with a multi-LED fusion strategy. The pinhole camera model is used to estimate the camera's position in the indoor environment. Over time, these solutions have been refined to achieve lower latency and higher throughput. For example, a low-latency algorithm specifically designed for rapidly-moving robots is proposed in~\cite{ebmer2024real}. Bias tuning is performed first, followed by frequency recognition for each LED. A low-pass filter supports tracking, and the PnP algorithm is applied to estimate the pose, by~mapping 3D LED positions to their 2D projections. In the same spirit, the PnP algorithm has been applied in~\cite{bauersfeld2025monocular} for the pose estimation of a drone. The authors proposed a pipeline achieving millimeter-level accuracy with a pose update rate up to 1~kHz and latency as low as 2.5~ms, using four IR~LEDs. This approach is particularly suited for small indoor environments and agile robots (e.g. mini quadrotors). 

EBOMS have also recently found application in \emph{aerospace robotics}. In~\cite{salah2023neuromorphic}, ground-level blinking LEDs are observed by an event camera mounted on a drone. A Kalman filter is designed for detecting and tracking the LEDs based on their unique frequencies. A real-time relative localization algorithm is then proposed, which relies on the Time-Delayed Kalman Filter (TDKF). The TDKF integrates pose estimates obtained by solving the PnP problem, with acceleration measurements from the onboard IMU (Inertial Measurement Unit). Running at 200~Hz, the algorithm in~\cite{salah2023neuromorphic} provides real-time pose estimates for the autonomous navigation of a ground and an aerial robot in planetary exploration missions.

In summary, the existing EBOMS-based pose estimation methods can be divided into three main families, each offering distinct advantages and each having inherent limitations. Frequency-coded multi-LED systems provide strong temporal signatures and are robust under fast motions. However, they rely on careful bias tuning and controlled lighting, and their accuracy drastically drops when only a few LEDs are visible. PnP-based pipelines with structured LED constellations achieve the highest spatial accuracy (millimeter-level accuracy at kilohertz update rates), yet they require stable geometric layouts and they are typically limited to small indoor environments where all LEDs are simultaneously visible. Finally, multi-modal approaches, that combine event data with an IMU or other sensors, offer greater robustness in highly-dynamic or outdoor environments, but introduce additional system complexity because of their dependence on auxiliary devices. In conclusion, no single family prevails over all others: each method excels in a specific regime determined by motion speed, lighting, workspace volume, and hardware constraints.


\subsection{Communication}\label{SectApp-comm}

Because of their unique properties, event cameras are well-suited for optical communication, especially in complex, unknown environments, where traditional frame-based cameras provide relatively poor results in terms of transmission speed and error~rate. As far as traditional optical communication systems are concerned, visible light communication (VLC) and IR-based devices are predominant for short-range data transmission~\cite{ghassemlooy2019optical}. These systems generally rely on photodiodes and frame-based cameras, as emitters and receivers, respectively~\cite{onodera2022drone}. However, they suffer from a number of weaknesses, such as a limited frame rate, motion blur in dynamic scenes, and sensitivity to ambient light~\cite{karunatilaka2015led,cahyadi2020optical}. Therefore, these systems are rarely used for high-speed applications in outdoor, dynamic environments.

Besides classical point-to-point communication, EBOMS could also be used in more challenging applications, such as inter-swarm communication, where multiple robots (securely) exchange messages via optical signals. This is particularly useful when radio-frequency communication is unreliable due to interference, jamming/spoofing, regulatory restrictions, or power constraints. In all these cases, optical emitters in conjunction with event-based receivers, provide a high-bandwidth and energy-efficient communication channel. 

In the remainder of this section, we present an overview of existing EBOMS for optical communication. We first analyze systems based on active optical markers such as LEDs, then consider geometric and passive markers, and finally describe hybrid solutions combining passive and active markers.

\begin{figure}[t!]
    \centering
    \includegraphics[width=0.94\linewidth]{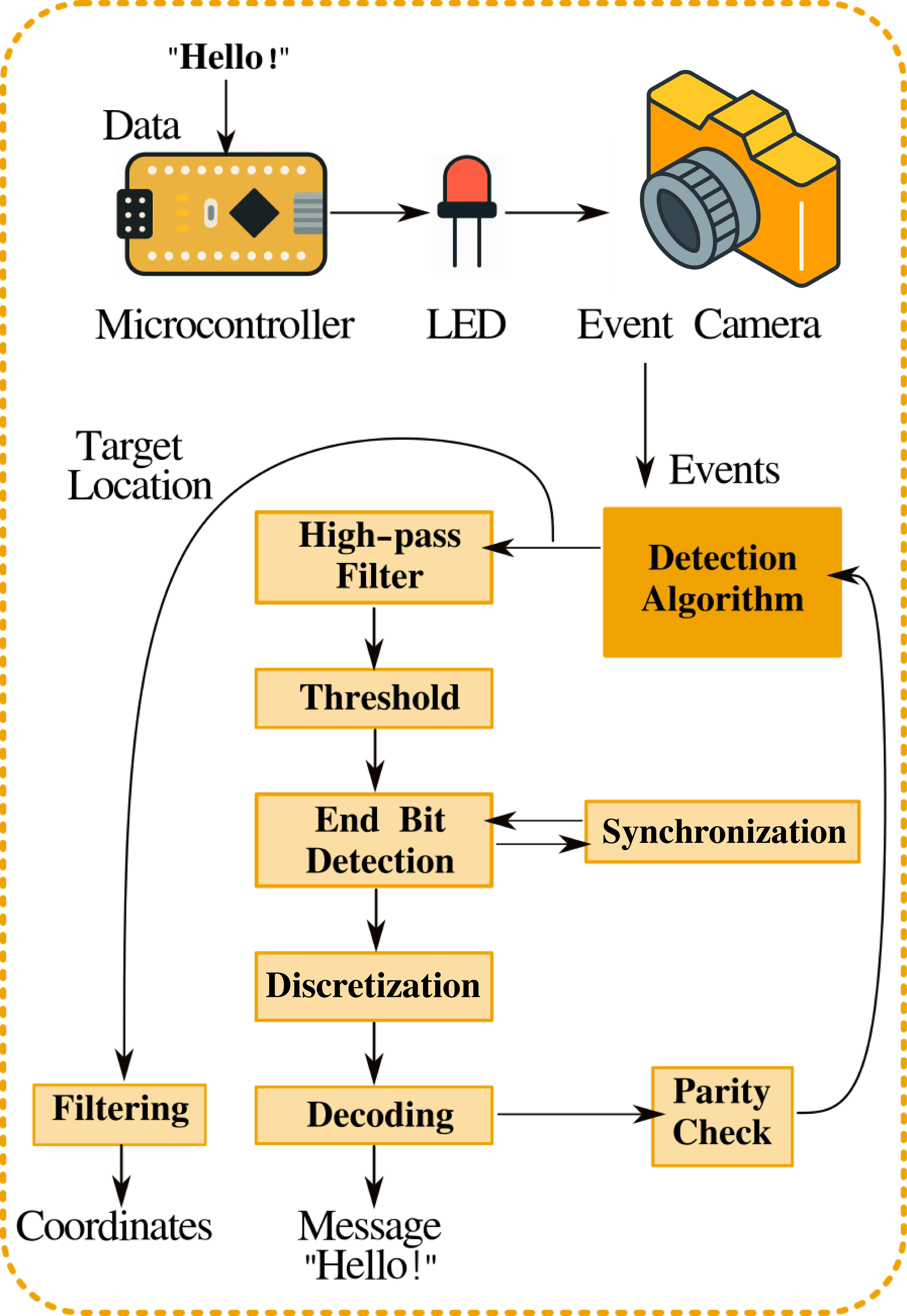}
    \caption{The EBOMS proposed in~\cite{wang2022smart} relies on LED-light encoding, LED~detection and asynchronous data decoding. The microcontroller modulates the light signals based on input data, which are then captured as events by the neuromorphic camera. The~detection algorithm processes these events, followed by asynchronous demodulation, decoding, and error checking, to~reconstruct the transmitted~message.}\label{fig: smart}
\end{figure}


\subsubsection{\textbf{Active markers}}

Various typologies of active markers, such as visible-light LEDs or IR LEDs, have been explored in the literature, each offering unique advantages depending on the specific application at hand. One of the foundational studies in this are is~\cite{censi2013low}. It laid the groundwork for further research, by showing that event cameras lend themselves to high-speed transmission via modulated LED signals, and by introducing a general framework for transforming raw event data into meaningful decoded messages. More specifically, in~\cite{censi2013low}, the authors proposed a low-latency algorithm to process event data and track IR LEDs blinking at different frequencies. They used evidence maps and a global particle filter, delivering superior performance compared to traditional cameras.

Building upon~\cite{censi2013low}, the authors in~\cite{vonarnim2024dynamic} developed a density-based clustering method to identify beacon signals from near-IR~LEDs. Event data is used to compute a sparse optical flow via spiking neural units, which estimate motion direction and speed. These estimates are then fed into a Kalman filter to track the beacon's position over time. During decoding, a start code is searched for, followed by message extraction and parity-bit checking for error detection. 

A significant departure from the state of the art is~\cite{wang2022smart}, where a smart EBOMS is presented (see Fig.~\ref{fig: smart} for the full pipeline). Messages are converted into compact \mbox{11-bit} binary packets: each packet includes a start code, a \mbox{7-bit} ASCII message, a~parity bit for error checking, and an end code. For the decoding process, a blob-detection algorithm is used to pinpoint the location of LED signals. A high-pass filter is then applied to extract the relevant high-frequency signals, and a thresholding technique allows to decode the original message. Thanks to the high-performance visible-light LEDs and special lenses for longer distances, error-free data transmission at a frequency of 500~bps over a distance of 100~m is achieved, even under bright sunlight. For the first time, this work demonstrated that active markers are effective for long-range optical communication in outdoor environments.

Recent research has further embraced the use of event cameras in the realm of optical communication. In~\cite{perez2019optical}, the authors introduced a high-speed camera-based communication system for data transmission via IR LEDs. The prototype also supports RGB LEDs in case of multicolor signaling. The system delivered excellent performances with low error rates at a distance of 5~m using a binary pulse-position modulation scheme with a specially-designed start-frame delimiter. Similarly, in~\cite{nakagawa2024linking}, the authors considered a multi-agent framework, which takes advantage of event-based visible-light communication. The proposed system guarantees an impressive success rate of 99.1\%, outperforming conventional radio-frequency and RGB-based  communication methods. It is robust to occlusions and motion blur, and the communication is reliable even at moderate distances and in visually-cluttered environments. 

In the last few years, efficient encoding schemes for optical communication have been studied. For example, in~\cite{aranda2024enhancing}, instead of using traditional methods to encode messages in LED-based systems (such as binary or ASCII code), a new approach, called \emph{$n$-pulse encoding}, is proposed. Information is conveyed by counting the number of pulses generated within a specific time window, which guarantees efficient data transmission. One of the key advantages of this solution is that it fully benefits from the low latency of event cameras and can effectively operate at high frequencies.

However, working with event cameras also poses unique challenges. For instance, for an object to be detected by an event camera, brightness changes are essential. To overcome this limitation, the authors in~\cite{tang2022preliminary} proposed a spinning array of LED transmitters. The after-image effect in conjunction with the spinning motion of the LEDs, ensures that the event camera perceives a continuous variation of brightness, without any loss of visibility. Although the rotating LED array uses the modulation of the blinking pattern based on the angular location of the LEDs to transmit the data, the decoding process is performed in two major steps. Noise is filtered by a sliding window, and then the signal is demodulated to accurately reconstruct the transmitted data. 

\subsubsection{\textbf{Geometric markers}}

LEDs are commonly employed as optical markers, but an interesting alternative, especially for communication purposes, is geometric markers (generally, ArUco markers). These fiducial markers have distinctive black-and-white square patterns, which can be easily detected and decoded via conventional computer-vision algorithms. Different supports are available for ArUco markers: they can be displayed on digital screens or printed on rigid substrates (in the latter case, an external source of light is necessary for visibility). Their flexibility and variety of form factors make them an excellent option in many applications.

In~\cite{su2024motion}, a novel transmission mechanism is proposed in which binary data is encoded as dynamic visual markers on a screen and captured by an event camera through asynchronous brightness changes. In order to ensure accurate spatial localization, the authors introduced the Surface of Active Events~(SAE). The~SAE allows the markers to maintain a high-contrast boundary and guarantees robust tracking. Data is transmitted as a video stream at 60~fps, and the decoding process involves binarizing the received information. Experimental results indicate that the system achieves a maximum transmission rate of 114~kbps under static conditions, with the rate decreasing to 7.68~kbps in case of motion.

\begin{figure}[b!]
    \centering
    \includegraphics[width=0.93\linewidth]{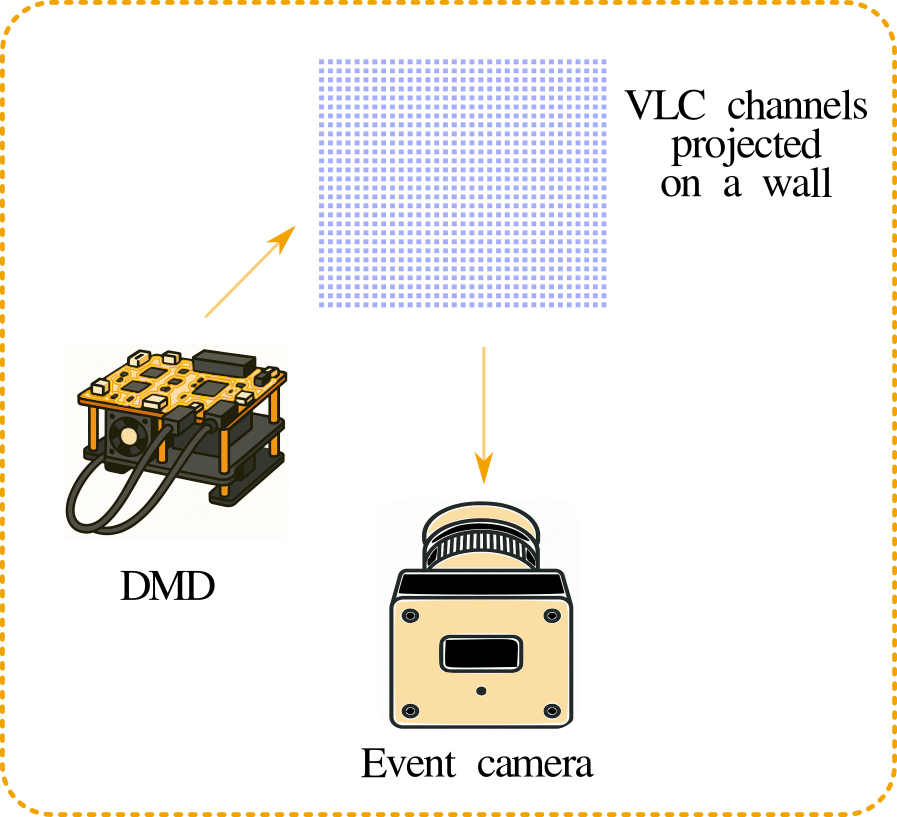}
    \caption{Overview of Selene EBOMS proposed in~\cite{wang2024towards}. The system consists of an event camera and a DMD (Digital Micro-mirror Device).}\label{fig: selene}
\end{figure}

A complementary approach is described in~\cite{sarmadi2021detection}. Here, the authors present a method for real-time decoding of binary square fiducial markers using an event camera. In a preprocessing step, a Gaussian filter is applied to the input data. The Line Segment Detector (LSD)~\cite{vonGioiJaMoRa_PAMI10} is then used to accurately identify the boundaries of the markers. The candidate marker regions are defined by matching line segments obtained after the temporal segmentation of event frames, which ensures a computationally-efficient detection pipeline. The binary message contained in the markers is retrieved by using a shift-convolution method applied to the marker grid and by taking advantage of the event distribution. 

\subsubsection{\textbf{Passive markers}}

Passive markers, a distinct class of optical markers, are widely used in EBOMS. They do not emit electromagnetic waves, but they instead reflect or modulate an external source of light. Reflective surfaces or Digital \mbox{Micro-mirror} Devices (DMDs) can be utilized as passive markers, especially in scenarios where adding an active light source is impractical due to environmental constraints or energy limitations. This class of optical markers is particularly useful when an obstacle lies between the event camera and the target: in this case, the neuromorphic sensor generates events by indirectly capturing the light incident on a reflective surface. Thus, the obstacle is detected, even if it is outside the \mbox{line of sight}.

Chief among the works in this area is~\cite{wang2024towards}, where the authors introduced Selene, a high-speed, multi-channel EBOMS based on passive optical communication principles. Thanks to the DMDs, no modulated light signals are emitted, and ambient illumination, such as sunlight, is sufficient (see Fig.~\ref{fig: selene}). First, a channel-mapping strategy clusters data into independent mirror-block positions. A duplicate event removal step filters out redundant events caused by strong intensity changes, retaining only the first polarity transition, that represents the actual state change of interest. Finally, a data-decoding step is carried out. A key innovation of Selene is the adoption of a dual refresh~rate: in fact, mirror blocks near the center refresh at a higher rate than those on the periphery. Since central pixels in event cameras exhibit lower timestamp latency than peripheral ones, the dual refresh rate significantly improves throughput without sacrificing decoding~accuracy. Experiments in~\cite{wang2024towards} demonstrate a peak data rate of 1.6~Mbps at distances up to 6~m, setting a new record for passive-marker-based communication. 

In a complementary approach presented in~\cite{nishar2025non}, the authors explore Non-Line-of-Sight (NLoS) communication using passive-light reflections from real-world objects. A modulated LED source and a neuromorphic camera are positioned orthogonally with respect to the target, so that only reflected light is captured. A periodic event-frame generation process is used to identify Regions of Interest (ROI), followed by contour-based refinement to extract the most informative pixels. Two modulation schemes are investigated: On-Off Keying (OOK) and an adaptive $n$-pulse encoding method that varies symbol mapping according to bit composition, to improve reliability. Demodulation is carried out using a sliding-window approach with guard intervals to address noise and multi-path interference. Experiments with various object surfaces show that glossy and reflective materials greatly enhance signal decoding. The~system achieves a bit error rate as low as $4 \times 10^{-4}$ in dark conditions, demonstrating the viability of passive markers for reliable NLoS communication. 

EBOMS based on passive markers have also been tested in challenging real-world applications, such as \emph{autonomous driving}. In~\cite{xu2023visible}, the authors introduced a visible-light communication system, called NeuromorphicVLC, which uses a retroreflective material and a liquid-crystal (LC) shutter to control reflected light with high precision. The detection algorithm starts with the creation of an evidence map to separate high-luminance areas. An adaptive quantizer encodes pixel values into three discrete values (On, Off, Noise): this allows to evaluate signal intensity and time consistency. Highly-reliable pixels are then chosen as key points, tracked over time and used in eight parallel channels to transmit messages. This~multichannel system is very accurate in dynamic conditions, but the operational distance is limited to about 2~m.

\begin{figure}[t!]
    \centering
    \includegraphics[width=0.88\linewidth]{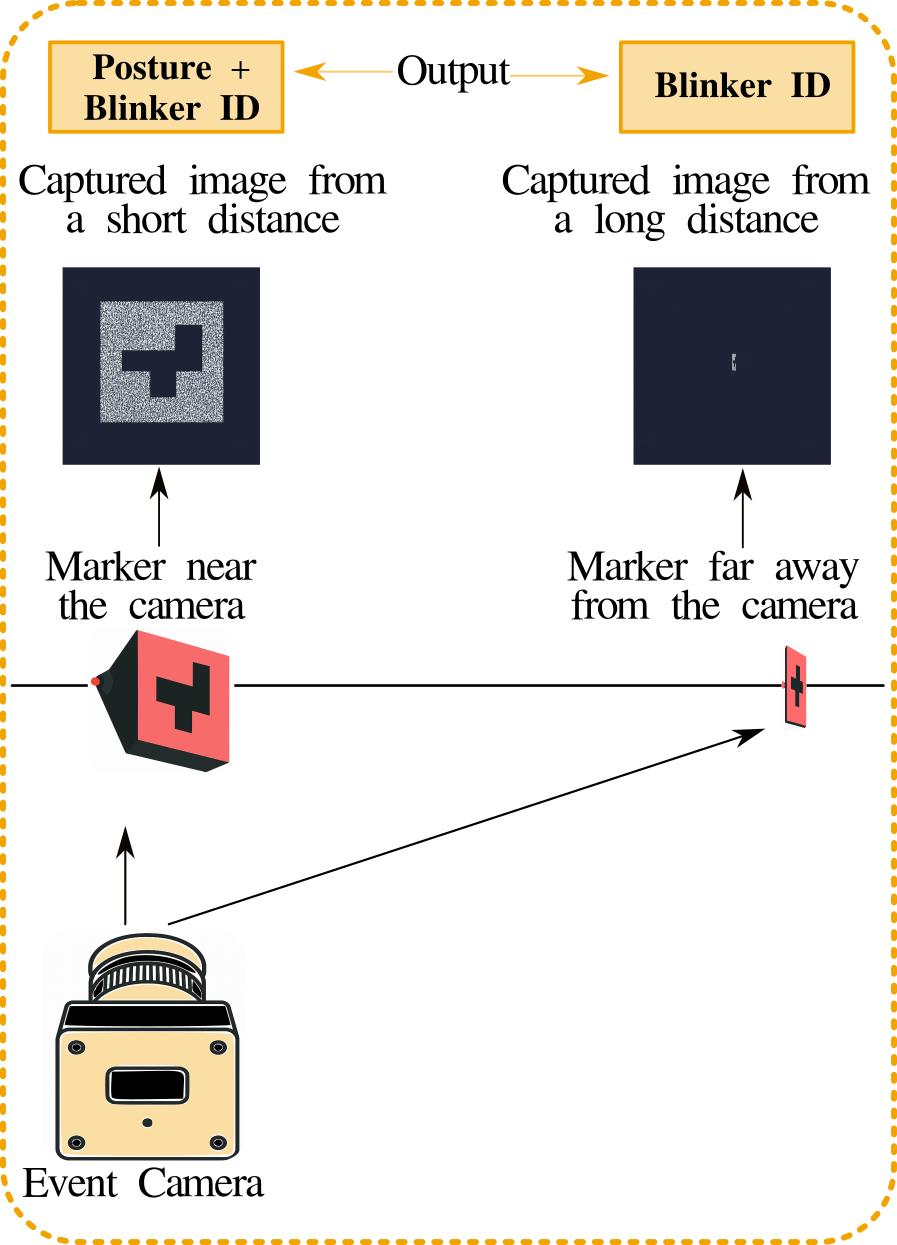}
    \caption{The Bicode system developed in~\cite{zhang2023bicode} is based on two modalities. It~can be used for pose estimation using geometric passive markers (close-range detection), but also for event-based optical communication using LEDs (long-range detection).}\label{fig: bicode}
\end{figure}

\subsubsection{\textbf{Hybrid markers}}

To enhance the accuracy and robustness of event-based optical communication systems, a final possible option consists in integrating different types of markers. A~representative example is the Bicode system~\cite{zhang2023bicode}, which aptly combines blinking IR LEDs with 2D markers to boost the signal-recognition rate. This hybrid design leverages the strengths of both modalities and hinges on advanced algorithms to process event data, enabling robust signal detection at distances up to 30~m. Moreover, as shown in Fig.~\ref{fig: bicode}, the geometric structure of the 2D markers also allows pose estimation, with objects as far as 2.5~m. One of the distinguishing features of the Bicode system is its transmission frequency of 571~bps, achieved with a low error rate.


\section{Conclusion and future research}\label{Sect:FutureWork}

This survey paper bridges a critical gap in knowledge and provides a systematic literature review of Event-Based Optical Marker Systems (EBOMS), over the last~15~years. We described the general working principles of EBOMS, examined the emerging technologies, and discussed the latest applications and best practices. We hope that this survey may serve as an entry point for practitioners and as a catalyst for research in this area at the crossroads of computer vision, pattern recognition and robotics.

Despite steady progress in the last decade, several open problems still exist in EBOMS, providing ample opportunity for future research. 

One of the most urgent needs is to develop \emph{standardized datasets and benchmarks}. In fact, the existing EBOMS are often evaluated in controlled laboratory settings, and there is a general lack of consistency in the environmental conditions and performance metrics. The field would greatly benefit from a unified, multi-modal dataset that includes synchronized event streams, IMU measurements and ground-truth pose estimates from an external motion-capture system, under diverse lighting and motion conditions. The availability of these resources is crucial to guarantee reproducibility, a fair comparative analysis, and the development of new data-driven methods. \mbox{The E-VLC} dataset~\cite{ShibaKoKo_CVPRW25a} is a welcome addition to the domain. It provides synchronized data from an event and a frame-based camera, and ground-truth pose annotations across diverse real-world conditions (indoor/outdoor, varying motion patterns, lighting levels, and camera settings). For the first time, thanks to the E-VLC dataset, the performance of new event-based communication and localization algorithms can be rigorously compared.

Equally important is the lack of \emph{standardized spatio-temporal encoding schemes} for event-based communication. Current systems rely on handcrafted or custom-designed modulation patterns, limiting scalability and interoperability. Future work should explore coding strategies which are adapted to the specificities of event cameras, such as spiking-based temporal encoding, frequency multiplexing with error correction, and more complex spatio-temporal marker layouts (see for example~\cite{HowardHi_ICCVW25}). Clearly, finding the right balance between communication speed, robustness, and energy efficiency, remains an open problem. 

Another key research direction pertains to the \emph{co-design of hardware and software} (detection algorithms). Most of existing EBOMS prototypes include off-the-shelf components and optimize software at a later stage. A tighter integration between marker design and asynchronous processing pipelines, would contribute towards higher accuracy and lower latency. For example, adaptive markers that respond to feedback from the environment (such as ambient light or detection confidence) by adjusting their emission level in real-time, could significantly enhance robustness.

Scaling up EBOMS to \emph{multi-agent systems} (e.g. swarms of drones) represents another major challenge. While a few works have explored inter-agent optical links, large-scale asynchronous communication in groups of autonomous vehicles, remains largely untapped. Future protocols could support event-based primitives that operate entirely without radio-frequency communication, e.g. for the identification of team members, optical token-passing, and agent coordination (e.g. formation control or cooperative target tracking).

\emph{Security and privacy} are also primary concerns. As EBOMS move from research laboratories to public spaces, they meet new challenges, such as spoofing and jamming attacks, data leakage and confidentiality issues. Research into secure marker encoding, such as frequency hopping, temporal steganography, or lightweight encryption combined with local edge-based decoding, could help developing privacy-aware systems for smart cities and safe human-robot interaction. 

We are also experiencing a surge in demand for \emph{miniaturized and application-specific EBOMS hardware}. Potential use cases include smart glasses (eye or head-worn wearable computers) that simultaneously estimate user's pose and communicate with their surroundings (e.g. for video games and AR~\cite{ShibaKoKo_CVPRW25b}), autonomous delivery robots moving in urban areas, and rovers exploring underground or underwater environments, where radio frequency-based localization is not reliable. In all these scenarios, form factor, power consumption, and resilience, are critical, and often conflicting, requirements.

On the marker-design front, \emph{more expressive and adaptable configurations} could be investigated. For example, flexible LED arrays could be wrapped around rigid or deformable objects and structured light emitted by DMDs or pulsed lasers could be used to project asynchronous patterns on static scenes~\cite{CaoLoLiMaZhPe_Measur26}. The latter solution would make passive communication or localization possible, without the need of physical markers attached to the objects. This idea has recently gained traction 
and it has been explored in~\cite{MorgensternGaBaHiEi_CVPRW23,SuSuWaXi_IROS25}, for depth estimation. The~development of advanced hybrid markers, combining both passive and active elements, can be regarded as a promising research direction. For example, systems that automatically switch between IR~LEDs and printed fiducials depending on the lighting conditions or power availability, would enjoy significant advantages in terms of accuracy and robustness. Moreover, multi-layered communication channels could be exploited by simultaneously relying on spatial, temporal, and frequency-based encoding.

Finally, one could take advantage of some well-known \emph{optical phenomena}, such as bokeh, for long-range EBOMS. In~fact, under optical defocus, bright markers, like LEDs, produce large, easily-detectable circular patterns on the images. These could be used to encode frequency, ID (identity), or motion information, even if an object is far way from the event camera. This seems particularly relevant for target detection, drone-to-drone communication, and cooperative robot exploration. 

In conclusion, research on EBOMS is still in its infancy but it has a bright future ahead, offering vast opportunities across hardware, software and algorithm design. To~face the challenges of tomorrow, researchers belonging to different communities (computer vision, robotics, machine learning, telecommunications, etc.) should join their efforts to usher in an era of intelligent and affordable event-driven perception~systems. 


\section*{Acknowledgments}

This work was supported by the French National Research Agency through the DEVIN project, ``\emph{Drones with Omni-Event Vision for Drone Neutralization}'' (ANR-23-IAS2-0001). The~authors would like to thank Dr. Fran\c{c}ois Rameau (SUNY Korea, South Korea) for fruitful discussions during the preparation of this survey paper.
	

\bibliographystyle{unsrt}
\bibliography{biblio_EventCam}

\end{document}

%% file: Tables/comparison_EBOMS-FBcameras.tex
\rowcolors{2}{gray!10}{white}
\definecolor{lightblue}{RGB}{239, 246, 255}
\rowcolors{2}{lightblue}{white}

\begin{table}[b!]
\begin{center}
\centering
\caption{Comparison of Event-Based Optical Marker Systems (EBOMS) and frame-based vision systems.}
\label{Table:Comparison}
\vspace{-0.175cm}
\begin{tabular}{p{1.6cm} p{2.72cm} p{3.3cm}}
\toprule
\textbf{Criterion} & \textbf{EBOMS} & \textbf{Frame-based vision systems} \\
\midrule
Latency & Microseconds ($\approx$ 10~$\mu$s) & Milliseconds (e.g., 16.7 ms at 60 fps) \\
Dynamic range & $>$ 120~dB (up to 140~dB) & Typically $\approx$ 60~dB \\ 
Motion blur & Negligible & Significant at high speed or with long \mbox{exposures} \\
Data type & Sparse, asynchronous & Dense, frames at fixed rate \\ 
\mbox{Power}  \mbox{consumption} & Sensor $+$ embedded \mbox{system} $\approx$ 0.5~W or~less & Typically between 1 and 5~W\\ 
Light conditions & Robust in a wide range of conditions & Susceptible to saturation\\ 
Spatial/temporal \mbox{resolution} & Low to medium spatial resolution, high temporal resolution & High spatial resolution (thanks to the consolidated CCD/CMOS technologies), low temporal resolution\\ 
Maturity \mbox{of algorithms} & Emerging, task-specific & Well-established ecosystem\\
\bottomrule
\end{tabular}
\end{center}
\end{table}

%

%% file: Tables/survey_papers.tex
\definecolor{DarkGreen}{rgb}{0,0.6,0}
\definecolor{DarkGray}{gray}{0.875}
\definecolor{LightYellow}{rgb}{1,1,0.8}
\rowcolors{2}{gray!10}{white}
\definecolor{lightblue}{RGB}{239, 246, 255}
\rowcolors{2}{lightblue}{white}
\newcommand{\yes}{\textcolor{DarkGreen}{\ding{51}}}
\newcommand{\no}{\textcolor{red}{\ding{55}}}

\begin{table*}[t!]
\centering
\caption{Overview of some representative EBOMS.}
\label{Table:Survey}
\resizebox{\textwidth}{!}{%
{ 
\fontsize{13}{14}\selectfont
\begin{tabular}{|c|c|c|c|P{2.9cm}|P{2.9cm}|P{2.2cm}|P{2cm}|P{4.7cm}|c|P{3.8cm}|P{3.5cm}|c|}
\hline
\rowcolor{DarkGray} 
\rotatebox{90}{\textbf{Reference}} & 
\rotatebox{90}{\textbf{Tracking}} & 
\rotatebox{90}{\textbf{Pose estimation}} & 
\rotatebox{90}{\textbf{Communication}  } & 
\rotatebox{90}{\textbf{Camera type}} {\rotatebox{90}{\textbf{(Manufacturer,}} \rotatebox{90}{\textbf{Model)}}} & 
\rotatebox{90}{\textbf{Marker type}} & 
\rotatebox{90}{\textbf{Frequency (Hz)}} & 
\rotatebox{90}{\textbf{Distance (m)}} & 
\rotatebox{90}{\textbf{Algorithms}} & 
\rotatebox{90}{\textbf{Error handling}} & 
\rotatebox{90}{\textbf{Indoor/}}
{\rotatebox{90}{\textbf{Outdoor (I/O),}}}
{\rotatebox{90}{\textbf{Lighting}}}
{\rotatebox{90}{\textbf{conditions}}} &  
\rotatebox{90}{\textbf{Target}} 
{\rotatebox{90}{\textbf{applications}}} & 
\rotatebox{90}{\textbf{Code available}} \\
\hline
\rowcolor{LightYellow} \multicolumn{13}{|c|}{\textbf{Object detection and tracking (Sect.~\ref{SectApp-ODT})}} \\
\hline
\cite{muller2011miniature} & \yes & \no & \no & iniVation\hspace{0.2cm} eDVS & LED & 500-1500 & 1-5 & Event counting, Time-difference tracking & \no & I/O, Natural light & Mobile robot tracking, pan-tilt tracking & \no \\
\cite{loch2023event} & \yes & \yes & \no & Prophesee\hspace{0.1cm} Gen3 EVK & Passive ArUco & N/A & Up to 1.75 & PnP, Event-based edge matching & \no & I, High contrast & Robot pose tracking & \no \\
\cite{zhang2023improved} & \yes & \no & \no & Prophesee EVK4 & Active ArUco & N/A & 0.3-0.5 & Mean shift, Corner refinement & \yes & I, Low light & Marker detection for robot vision & \no \\
\cite{huang2023novel} & \yes & \yes & \no & Prophesee EVK4 & Active ArUco & N/A & 0.3-0.5 & Harris corner detection, Mean shift & \no & I, Poor exposure & Pose estimation for robotics & \no \\
\hline
\rowcolor{LightYellow} \multicolumn{13}{|c|}{\textbf{Pose estimation (Sect.~\ref{SectApp-PE})}}\\
\hline
\cite{xu2019method} & \no & \yes & \no & iniVation eDVS-4337 & Infrared LED & 1000 & 1.6-3.2 & Clustering, PnP, Least-squares pose optimization & \no & I, Low and high light & Pose tracking for drones & \no\\
\cite{chen2020novel} & \no & \yes & \no & iniVation DAVIS346Red & Infrared LED & 300-1500 & 0.4-2 & GM-PHD filter, Frequency detection & \yes & I & Indoor positioning system & \no \\ 
\cite{ebmer2024real} & \yes & \yes & \no & Prophesee EVK4 IMX636ES & Infrared LED & 4000-40000 & 2.1-4.8 & Bias tuning, PnP & \no & \hspace{0.5cm}I, Low light\newline O, Sunlight & 6-DoF pose estimation for drones & \yes \\
\cite{bauersfeld2025monocular} & \no & \yes & \no & Prophesee\hspace{0.1cm} Gen3 & Infrared LED & 1730-3520 & 0.7-5 & Signed Delta-Time Volume, Frequency Matching, PnP & \no & I, Bright light & Monocular motion capture for quadrotors & \no \\
\cite{salah2023neuromorphic} & \no & \yes & \no & iniVation DVXplorer & Infrared LED & 200-600 & Up to 7 & Clustering based on Gaussian mixture models, TDKF & \no & I & Relative localization for space robotics & \no \\
\cite{censi2013low} & \no & \yes & \yes & iniVation DVS128 & Infrared LED & 1000-2000 & 1-5 & Particle filter, Frequency clustering & \no & I, Low light & Pose tracking and communication for drones & \yes \\
\hline
\rowcolor{LightYellow} \multicolumn{13}{|c|}{\textbf{Communication (Sect.~\ref{SectApp-comm})}}\\
\hline
\cite{vonarnim2024dynamic} & \yes & \no & \yes & iniVation DVXplorer Mini & Infrared LED & 50-5000 & 0.5-28 & Density clustering, Spiking NN, Kalman filter & \yes & I/O, Sunlight & Tracking and communication for drones & \yes \\
\cite{wang2022smart} & \yes & \no & \yes & Prophesee\hspace{0.1cm} Gen3 & Visible-light LED & 500 & 1-100 & Blob detection, High-pass filtering & \yes & I/O, Sunlight & Beacon-based communication for robotics & \yes \\
\cite{perez2019optical} & \no & \no & \yes & N/A & \hspace{0.3cm}Infrared,\newline RGB LED & 500 & Up to 5 & Cluster detection, Matched filtering & \yes & I & High-speed optical communication & \no \\
\cite{nakagawa2024linking} & \yes & \no & \yes & Prophesee\hspace{0.1cm} IMX636 & Visible-light LED & 250 & Up to 5.5 & Event-based VLC decoding, Binary encoding & \no & I/O, Sunlight & Multi-agent communication & \no \\
\cite{aranda2024enhancing} & \no & \no & \yes & iniVation DVXplorer & Visible-light LED & 200 & 0.2 & $n$-pulse, Event counting demodulation & \no & I, Dark room & Low-complexity VLC & \no \\
\cite{tang2022preliminary} & \no & \no & \yes & iniVation DVXplorer Lite & Visible-light LED & N/A & 1 & Sliding window denoising, After-image demodulation & \no & I & Event-based VLC with rotating LED & \no \\
\cite{wang2024towards} & \no & \no & \yes & iniVation DVXplorer Lite & Passive DMD & Up to 2000 & Up to 6 & Channel mapping, Duplicate event removal & \yes & I, from 0 to 433 lux & High-speed passive VLC & \no \\
\cite{xu2023visible} & \yes & \no & \yes & Virtual Spike Camera & Passive LC shutter & Up to 6400 & 0.5-2, Up~to 800 (simul.) & Adaptive quantization, Key point tracking & \no & I/O, Sunlight & Blind-spot and road condition alerts & \no \\
\cite{nishar2025non} & \yes & \no & \yes & Prophesee EVK4 & Passive object reflection & Up to 1000 & 1 & ROI refinement, Adaptive $n$-pulse & \yes & \hspace{0.45cm}I, Dark room,\newline ambient light & Static NLoS communication & \no \\
\cite{su2024motion} & \yes & \yes & \yes & Prophesee\hspace{0.1cm} Gen4 & Digital marker & 60 & 1.25 & Refractory filtering, SAE normalization & \no & \hspace{0.44cm}I, Dark, normal\newline and sunlight & Optical communication, AR localization & \yes \\
\cite{sarmadi2021detection} & \yes & \yes & \yes & iniVation DVS128 & Passive ArUco & N/A & 0.2-0.3 & LSD, Gaussian convolution & \yes & I & Pose estimation, and optical communication & \yes \\
\cite{zhang2023bicode} & \yes & \yes & \yes & Prophesee EVK4 & Hybrid & 1000 & 2.5-20 & SORT algorithm~\cite{BewleyGeOtRaUp_ICIP16}, Manchester coding & \yes & I/O, Evening light & Hybrid tracking, localization & \no \\
\hline
\end{tabular}
}
}
\end{table*}